\documentclass[runningheads]{llncs}
\usepackage{graphicx}
\usepackage{todonotes}
\usepackage{multirow}

\usepackage{array}
\usepackage{tabu}
\newcolumntype{M}[1]{>{\centering\arraybackslash}m{#1}}
\newcolumntype{L}[1]{>{\arraybackslash}m{#1}}

\begin{document}
\title{Overview of BioASQ 2024: The twelfth BioASQ challenge on Large-Scale Biomedical Semantic Indexing and Question Answering
}
    \titlerunning{Overview of BioASQ 2024}
%

\author{
Anastasios Nentidis\inst{1,2} \and
Georgios Katsimpras\inst{1} \and
Anastasia Krithara\inst{1} \and 
Salvador Lima-López\inst{3} \and 
Eulàlia Farré-Maduell\inst{3} \and
Martin Krallinger\inst{3} \and
Natalia Loukachevitch\inst{4} \and
Vera Davydova\inst{5} \and
Elena Tutubalina\inst{5,6,7} \and
Georgios Paliouras\inst{1}
}
\authorrunning{A. Nentidis et al.}
%
\institute{
National Center for Scientific Research ``Demokritos'', Athens, Greece\\
\email{\{tasosnent, gkatsibras, akrithara, paliourg\}@iit.demokritos.gr}\\
\and
Aristotle University of Thessaloniki, Thessaloniki, Greece\\ 
\and
Barcelona Supercomputing Center, Barcelona, Spain\\
\email{\{slimalop, efarre, martin.krallinger\}@bsc.es}\\
\and
Moscow State University, Russia 
\and
Sber AI, Russia 
\and
Artificial Intelligence Research Institute, Russia 
\and
Kazan Federal University, Russia\\
\email{tutubalinaev@gmail.com}
}
\maketitle              
\begin{abstract}
This is an overview of the twelfth edition of the BioASQ challenge in the context of the Conference and Labs of the Evaluation Forum (CLEF) 2024. 
BioASQ is a series of international challenges promoting advances in large-scale biomedical semantic indexing and question answering. 
This year, BioASQ consisted of new editions of the two established tasks b and Synergy, and two new tasks: a) MultiCardioNER on the adaptation of clinical entity detection to the cardiology domain in a multilingual setting, and b) BIONNE on nested NER in Russian and English.
In this edition of BioASQ, 37 competing teams participated with more than 700 distinct submissions in total for the four different shared tasks of the challenge. 
Similarly to previous editions, most of the participating systems achieved competitive performance, suggesting the continuous advancement of the state-of-the-art in the field.  

\end{abstract}
\section{Introduction}
The BioASQ challenge has been focusing on the advancement of the state-of-the-art in large-scale biomedical semantic indexing and question answering (QA) for more than 10 years~\cite{Tsatsaronis2015}. 
To this end, it organizes different shared tasks annually, developing respective benchmark datasets that represent the real information needs of experts in the biomedical domain. 
This allows the participating teams from around the world, who work on the development of systems for biomedical semantic indexing and question answering, to benefit from the publicly available datasets, evaluation infrastructure, and exchange of ideas in the context of the BioASQ challenge and workshop. 

Here, we present the shared tasks and the datasets of the twelfth BioASQ challenge in 2024, as well as an overview of the participating systems and their performance.
The remainder of this paper is organized as follows. 
First, Section~\ref{sec:tasks} presents a general description of the shared tasks, which took place from January to May 2024, and the corresponding datasets developed for the challenge. 
Then, Section~\ref{sec:participants} provides a brief overview of the participating systems for the different tasks. 
Detailed descriptions for some of the systems are available in the proceedings of the lab. 
Subsequently, in Section~\ref{sec:results}, we present the performance of the systems for each task, based on state-of-the-art evaluation measures or manual assessment.
Finally, in Section~\ref{sec:conclusion} we draw some conclusions regarding the 2024 edition of the BioASQ challenge.

\section{Overview of the tasks}
\label{sec:tasks}

The twelfth edition of the BioASQ challenge consisted of four tasks:
(1) a biomedical question answering task (task b), (2) a task on biomedical question answering for open developing issues (task Synergy), both tasks considering documents in English, (3) a new task focused on the automatic detection of disease and drug mentions (task MultiCardioNER), considering cardiology clinical case documents in Spanish, English, and Italian, and (4) a new task on NLP challenges on biomedical nested named entity recognition (NER) systems for English and Russian languages (task BIONNE)~\cite{nentidis2024bioasq}.
In this section, we first describe this year's editions of the two established tasks b (task 12b) and Synergy (Synergy 12)~\cite{BioASQ2024task12bSynergy} with a focus on differences from previous editions of the challenge~\cite{nentidis2023overview,nentidis2022overview}. Additionally, we also present the new tasks MultiCardioNER on multiple clinical entity detection in multilingual medical content~\cite{multicardioner2024overview}, and BIONNE on nested NER in Russian and English~\cite{bionne}.

\subsection{Task 12b}
The twelfth edition of task b (task 12b) focuses on a large-scale question-answering scenario in which the participants are required to develop systems for all the stages of biomedical question answering. 
As in previous editions, the task examines four types of questions: “yes/no”, “factoid”, “list” and “summary” questions \cite{balikas13}.
In this edition, the training dataset provided to the participating teams for the development of their systems consisted of 5,049 biomedical questions from previous versions of the challenge annotated with ground-truth relevant material, that is, articles, snippets, and answers~\cite{krithara2023bioasq,BioASQ2024task12bSynergy}. 
Table \ref{tab:b_data} shows the details of both training and test datasets for task 12b.
The test data for task 12b were split into four independent bi-weekly batches consisting of 85 questions each, as presented in Table \ref{tab:b_data}. 

\begin{table}[!htb]
        \caption{Statistics on the training and test datasets of task 12b. The numbers for the documents and snippets refer to averages per question.}\label{tab:b_data}
        \centering
        \begin{tabular}{M{0.09\linewidth}M{0.08\linewidth}M{0.08\linewidth}M{0.09\linewidth}M{0.15\linewidth}M{0.15\linewidth}M{0.15\linewidth}M{0.15\linewidth}}\hline
        \textbf{Batch} 	& \textbf{Size} 	&	\textbf{Yes/No}	&\textbf{List}	&\textbf{Factoid}	&\textbf{Summary}& \textbf{Documents} 	& \textbf{Snippets}  	\\ \hline
        Train   & 5,049 & 1,357 & 967 & 1,515 & 1,210 & 9.06 & 11.91 \\
        Test 1	& 85   & 25   & 21  & 21   & 18   & 3.20 & 4.36  \\
        Test 2	& 85   & 26   & 18  & 19   & 22   & 2.72 & 3.69  \\
        Test 3	& 85   & 24   & 19  & 26   & 16   & 2.45 & 3.36  \\
        Test 4	& 85   & 27   & 22  & 19   & 17   & 2.18 & 3.44  \\\hline  
        \textbf{Total} & 5,389 & 1,459 & 1,047 & 1,600 & 1,283 & 8.65 & 11.4  \\\hline 
        \end{tabular}
\end{table}

As in previous editions of task b, task 12b was also divided into distinct phases. Contrary to previous editions, however, task 12b was divided into three phases:
a) In phase A, a test set consisting of the bodies of biomedical questions, written in English, was released and the participants had 24 hours to identify and submit relevant PubMed/MEDLINE-article abstracts, and snippets extracted from them. 
b) In phase A+, which runs in parallel to phase A, the participants could also submit \textit{exact answers}, that is entity names or short phrases, and \textit{ideal answers}, that is, natural language summaries of the requested information for the same questions.
c) In phase B, which runs after the completion of phases A and A+, some relevant articles and snippets were released for these questions, and the participating systems had another 24 hours to respond with \textit{exact} and \textit{ideal answers} taking this manually selected material into account. 

For example, for the “yes/no” question "Is levosimendan effective for amyotrophic lateral sclerosis?", the systems in phase A should respond with relevant documents and snippets useful for providing an answer. In parallel, systems participating in phase A+ could also attempt to provide the \textit{exact} and \textit{ideal answer}, which are "No." and "No. Levosimendan was not superior to placebo in maintaining respiratory function in a broad population with amyotrophic lateral sclerosis. Although levosimendan was generally well tolerated, increased heart rate and headache occurred more frequently with levosimendan than with placebo." respectively. In Phase B, sufficient relevant documents and snippets were also released, and the participating systems had to return \textit{exact} and \textit{ideal answer} again, exploiting this information.

\subsection{Task Synergy 12}
The task Synergy was introduced three years ago~\cite{nentidis2021overview} envisioning a continuous dialog between the experts and the automated question-answering systems. 
In this model, the systems provide relevant material and answers to the experts who posed some open questions for developing problems. The experts assess these responses and feed their assessment back to the systems, including the information on whether the material retrieved is sufficient for providing an answer to this question (``answer ready''). 
This feedback is then exploited by the systems together with new material, that becomes available in the meantime, to provide updated responses to the experts.
This process proceeds with new feedback and new responses for the same open questions that persist, in an iterative way, organized in rounds. 

After twelve rounds in the context of BioASQ9~\cite{krithara2021bioasq} and BioASQ10~\cite{nentidis2022bioasq}, focusing on open questions about the COVID-19 pandemic, in BioASQ11 we extended the Synergy task with four rounds open to any developing problem~\cite{nentidis2023bioasq}. In BioASQ12, we continue in this setting with four more bi-weekly rounds open to any developing problem of interest for the six biomedical experts who participated this year.
As in previous versions, the open questions were not required to have definite answers, and a distinct version of PubMed/MEDLINE was designated per round for relevant material retrieval.
A set of 311 questions with respective incremental expert feedback and answers from the previous versions of the task, was available as a development set. 
This year, 73 distinct questions were used in the new rounds of the task. Of them, 18 questions were persisting from the previous versions and 55 were new.  
The distribution of the Synergy 12 questions round is shown in Table \ref{tab:syn_data}.
 
\begin{table}[!htb]
	\caption{Distribution of the questions of Task Synergy 12 per round.
 }\label{tab:syn_data}
    \centering
    \begin{tabular}{c c c c c c c c}\hline
     \textbf{Round}  & \textbf{Size}  & \textbf{Yes/No} &\textbf{List} &\textbf{Factoid} &\textbf{Summary}& \textbf{Answer  ready}  \\
    \hline
         1  & 72 & 11 & 29 & 17 & 15 & 33 \\
         2  & 72 & 11 & 29 & 18 & 14 & 46 \\
         3  & 64 & 10 & 24 & 16 & 14 & 50 \\
         4  & 64 & 10 & 24 & 17 & 13 & 57 \\
     \hline                 
    \end{tabular}
\end{table}
 
 The same types of questions (yes/no, factoid, list, and summary) and answers (\textit{exact} and \textit{ideal}) are examined in this task as in task 12b, and the same evaluation measures are adopted for system assessment. 
 However, task Synergy is not structured into phases, with both relevant material and answers received together. 
 For new questions, only relevant material was required until the expert marked a question as ``answer ready''. Then, both new relevant material and answers are expected for it in subsequent rounds. In case a question receives a definite answer that is not expected to change, the expert can mark it as ``closed'' to be excluded from the remaining rounds.

\subsection{Task MultiCardioNER}
The MultiCardioNER shared task is a continuation of previous tasks focused on the detection of named entities in clinical case reports in Spanish such as DisTEMIST \cite{amiranda2022overview} for diseases, MedProcNER/ProcTEMIST \cite{medprocner} for procedures and SympTEMIST \cite{symptemist} for signs and symptoms. 
While these previous tasks use a general collection of texts from multiple clinical specialties, MultiCardioNER focuses on the creation of systems that can detect diseases and drugs specifically in the cardiology domain. For this purpose, in addition to the DisTEMIST and DrugTEMIST corpus (which include disease and drug annotations, respectively, for the same collection of varied clinical case reports), participants were provided the CardioCCC corpus of cardiological documents. This new dataset includes 508 documents, out of which 250 were reserved to be used as test set and the rest were released so that participants could use them as they saw fit. 
As an added novelty, this task introduces a multilingual aspect with the release of the Gold Standard drug annotations in English and Italian, as well as in Spanish.
Table \ref{tab:multicardio_corpus-overview} presents a summary of the different corpora used as part of the task.

\begingroup
\setlength{\tabcolsep}{4pt}
\begin{table}[]
    \centering
    \caption{Statistics for the datasets provided for MultiCardioNER. ``Annot.'' stands for ``annotations'', while ``Chars'' stands for ``characters''.
    Unique annotations refer to the number of distinct annotated strings after converting all annotations to lowercase.
    The number of tokens has been calculated using the following spaCy models: ``es\_core\_news\_sm'', ``en\_core\_web\_sm'' and ``it\_core\_news\_sm''.}
    \begin{tabular}{lccccccM{0.12\linewidth}}
        \hline
        \textbf{Dataset} & \textbf{Lang.} & \textbf{Entity} & \textbf{Docs} & \textbf{Tokens} & \textbf{Chars} & \textbf{Annot.} & \textbf{Unique Annot.} \\
        \hline
        DisTEMIST & ES & Diseases & 1,000 & 406,137 & 2,335,968 & 10,664 & 6,739 \\
        DrugTEMIST & ES & Drugs & 1,000 & 406,137 & 2,335,968 & 2,778 & 925 \\
         & EN & Drugs & 1,000 & 404,194 & 2,230,631 & 2,814 & 875 \\
         & IT & Drugs & 1,000 & 421,251 & 2,393,002 & 2,808 & 893 \\
        \hline
        CardioCCC & ES & Diseases & 508 & 568,297 & 3,215,774 & 18,232 & 7,692 \\
         & ES & Drugs & 508 & 568,297 & 3,215,774 & 4,227 & 755 \\
         & EN & Drugs & 508 & 576,772 & 3,114,833 & 4,231 & 734 \\
         & IT & Drugs & 508 & 595,332 & 3,345,466 & 4,385 & 752 \\
        \hline
    \end{tabular}
    \label{tab:multicardio_corpus-overview}
\end{table}
\endgroup

All in all, MultiCardioNER is divided into two different subtracks:

\begin{itemize}
    \item \textbf{Subtrack 1 (CardioDis)}. This track focuses on the adaptation of disease recognition systems to the cardiology domain in Spanish. Some examples of cardiology-specific diseases would be ``atrial flutter with rapid ventricular response'' or ``Takotsubo syndrome''. Participants were provided the DisTEMIST corpus \cite{amiranda2022overview}, as well as a new collection of 258 cardiology-specific clinical case reports annotated with diseases (CardioCCC). They were allowed to distribute the data collections however they saw fit in order to achieve the best system possible. The evaluation was done in the second half of the CardioCCC corpus, made up of 250 documents, using strict, micro-averaged precision, recall and F1-score. The annotation for these documents was done following the DisTEMIST annotation guidelines, which are available on Zenodo\footnote{\url{https://zenodo.org/doi/10.5281/zenodo.6458078}}
    \item \textbf{Subtrack 2 (MultiDrug)}. This track focuses on the multilingual (Spanish, English and Italian) adaptation of medication recognition systems to the cardiology domain. Some examples of medication entities are ``nytroglicerine'' and ``clopidogrel''. For this track, participants were provided the DrugTEMIST dataset, which is a companion corpus to the previously-released DisTEMIST, ProcTEMIST and SympTEMIST corpora that offer annotations of medications for the same collection of texts and hadn't been released until now. As in the previous track, a portion of the cardiology-specific dataset CardioCCC was released during the training phase and another was reserved to be used as test set. While the original versions of both datasets were created using Spanish texts, a machine-translated version in English and Italian was revised and annotated by clinical experts native in each language. These documents were annotated following the DrugTEMIST guidelines, published specifically for this task and available on Zenodo\footnote{\url{https://zenodo.org/doi/10.5281/zenodo.11065432}}. The evaluation for this subtrack was also done using strict, micro-averaged precision, recall and F1-score, with every language being evaluated separately.
\end{itemize}

The MultiCardioNER datasets are publicly available to download on Zenodo\footnote{\url{https://zenodo.org/doi/10.5281/zenodo.10948354}}. In addition to the Gold Standard datasets, a background set of related clinical case reports (including both cardiology and non-cardiology documents) was also released as part of the task. This background set includes 7,625 documents and is available in Spanish, English and Italian. There are some documents originally being written in each language, with the rest being translated via machine translation. A Silver Standard that aggregates the predictions of the participant systems for this documents will also be released and uploaded to the same repository.

MultiCardioNER is promoted by Spanish and European projects such as DataTools4Heart, AI4HF, BARITONE and AI4ProfHealth and organized by the Barcelona Supercomputing Center (BSC) in collaboration with BioASQ. A more in-depth analysis of the MultiCardioNER Gold Standard, guidelines and additional resources is presented in the MultiCardioNER overview paper \cite{multicardioner2024overview}.

\subsection{Task BioNNE}
Given that most biomedical datasets and named entity recognition (NER) methods are designed to identify flat, non-nested mention structures, we introduce the Biomedical Nested Named Entity Recognition (BioNNE) shared task this year. For instance, in the text ``[[[eye] movement] disorders]'', nested annotations assist in identifying and distinguishing both the broader categories, like medical conditions and physiological functions, as well as the specific anatomical parts involved.
The main task focuses on extracting and classifying biomedical nested named entities from unstructured PubMed abstracts available in both Russian and English. It is divided into three tracks:
\begin{itemize}
    \item \textbf{Bilingual}: Participants develop a single multilingual NER model using data in both Russian and English, generating predictions for each language.
    \item \textbf{English-oriented or Russian-oriented}: Participants build a nested NER model specifically for abstracts in one target language, either English or Russian.
\end{itemize}
The training and validation sets for the BioNNE competition were derived from a subset of the NEREL-BIO dataset \cite{NERELBIO}. This dataset enhances the original NEREL \cite{loukachevitch2023nerel} dataset, which was designed for the general domain by incorporating biomedical entity types. We made improvements by correcting annotator errors, merging the \texttt{PRODUCT} and \texttt{DEVICE} classes into a unified \texttt{DEVICE} class, and selecting the eight most frequent medical entities: \texttt{FINDING}, \texttt{DISO}, \texttt{INJURY\_POISONING}, \texttt{PHYS}, \texttt{DEVICE}, \texttt{LABPROC}, \texttt{ANATOMY}, and \texttt{CHEM}. The final dataset includes 662 annotated PubMed abstracts in Russian and 104 parallel abstracts in both Russian and English. In total, there are 40,782 annotated entities in Russian and 8,099 in English.
A new test set was developed specifically for the shared task, including 154 abstracts in both English and Russian, each containing $\approx$10k annotated entities.

All the materials can be found on BioNNE GitHub page\footnote{\url{https://github.com/nerel-ds/NEREL-BIO/tree/master/bio-nne/}} and on CodaLab\footnote{\url{https://codalab.lisn.upsaclay.fr/competitions/16464}} competition page.

\section{Overview of participation}
\label{sec:participants}
\subsection{Task 12b}

This year, 26 teams participated in task 12b, submitting a total of 89 different systems across all three phases A, A+, and B. Specifically, 18, 8, and 16 teams competed in phases A, A+, and B, with 64, 34, and 54 distinct systems respectively. Eight of these teams were involved in all three phases.
An overview of the technologies utilized by the teams is outlined in Table \ref{tab:b_sys}. Additional details for specific systems can be found in the workshop proceedings. 
As in previous years, the open-source system OAQA~\cite{yang2016learning}, which achieved top performance in older editions of BioASQ, was used as a baseline for phase B \textit{exact answers}. This system is based on the UIMA framework and relies on traditional NLP and Machine Learning approaches and tools, such as MetaMap and LingPipe~\cite{baldwin2003lingpipe}.

\begin{table}[!htb]
        \centering
         \caption{Systems and approaches for task 12b. Systems for which no information was available at the time of writing are omitted.}
        \begin{tabular}{M{0.23\linewidth}M{0.1\linewidth}M{0.05\linewidth}M{0.56\linewidth}}\hline
        \textbf{Systems} & \textbf{Phase}& \textbf{Ref.}& \textbf{Approach} \\ \hline
        MQU	 & A,A+,B & \cite{Galat24clef} & llama-2, llama-3, gemini, phi-3, query expansion, re-ranking, RAG  \\\hline 
        BSRC & A,A+,B & \cite{Panou24} & open source LLMs, sparse/dense/hybrid retrieval  \\\hline
        MiBi & A,A+,B & \cite{Reimer24} & BM25, re-ranking, RAG, GPT-3.5, GPT-4, Mixtral-8x7B, DSPy LLM\\\hline 
        UR	 & A,A+,B & \cite{Ateia24clef} & Claude 3 Opus, GPT-3.5-turbo, Mixtral 8x7B, adapter fine-tuning, query expansion \\\hline %
        UA	 & A,A+,B & \cite{Almeida24clef} & BM25, PubMedBERT, BioLinkBERT, llama-2, llama-3, Nous-Hermes2-Mixtral, Gemma-2b, \\\hline
        CUHK-AIH & A,A+,B & \cite{Gao24clef} & BM25, llama-2, RAG  \\\hline
        Gatech & A,A+,B & \cite{Zhou24clef} &  Mixtral, GPT-J, GPT-4, Llama2, resampling\\\hline
        Fudan-Atypon & A,A+,B & - &  GPT3.5/4.0, ChatGLM, Spark, scenario prompt, LLama3-8B-instruct, query expansion \\\hline
        UNIPD & A  & \cite{Huang24clef}&  BM25, BiomedBERT, GPT-3.5, Gemini, NER \\\hline %
        OPIX & A  & - & BM25, GAT, re-ranking, DRUMS \\\hline %
        HU & A  & \cite{Şerbetçi24clef} & BM25, MedCPT, E5,  GPT-3.5 \\\hline %

        NCU & B & \cite{Chih24clef}& GPT-4, RAG\\\hline  %
        UL & B & \cite{Anaya24} & Mistral-7B-instruct, iterative fine-tuning \\\hline %
        VCU & B & - & Synthia-13B, llama-3 \\\hline  

        \end{tabular}
       \label{tab:b_sys}
\end{table}

The MQU team from Macquarie University participated in all three phases of the task with five systems. Their systems relied on several open Large Language Models (LLMs) like llama-2, llama-3, gemini, and phi-3. Additionally, the team applied techniques like query expansion, re-ranking, and Retrieval Augmented Generation (RAG) on abstracts to improve their results.
Another team participating in all phases is the team from the BSRC Alexander Fleming Institute. Their systems focused on sparse, dense and hybrid methods for document and snippet retrieval and LLMs with optimized prompts for exact and ideal answers. For document and query embedding as well as answer generation open LLMs were employed. Additionally, for Yes/No questions, a jury of complementary open LLMs perform majority voting to determine the final results.

The MiBi team from the Friedrich-Schiller-Universität Jena also participated in all phases with five systems. Their systems relied on LLM-based RAG. In phase A, their systems applied BM25 scoring to the article’s title, abstract, and MeSH terms, while snippets were extracted either by GPT-3.5 chain-of-thought few-shot prompting or heuristically by re-ranking the title and chunks of up to three sentences from the abstract. They also performed re-ranking using pre-trained bi-encoders, cross-encoders, and lexical BM25 scoring. For phase A+, their systems were based on zero-shot prompting with GPT-3.5, GPT-4, or Mixtral-8x7B. The team also used the DSPy LLM programming framework for some runs. Furthermore, in phase B, their systems relied either on prompting an LLM (GPT-3.5 and GPT-4) with top-3 ``ground truth" abstracts or all the ``ground truth" snippets as context.

The UR team from the Universität Regensburg competed in all phases of the task with five systems. Their systems employed 1-shot and 10-shot learning combined with adapter fine-tuning for both commercial (Claude 3 Opus, GPT-3.5-turbo) and open source models (Mixtral 8x7B). In phase A, they relied on different 1-shot and 10-shot learning  settings including plain, fine-tuned or models with additional context retrieved from Wikipedia, while in phases A+ and B, their systems relied  solely on 10-shot learning.

The UA team from the Uni de Aveiro participated in all three phases of the task with five systems. In phase A, their systems followed a two-stage retrieval pipeline for the document retrieval using the traditional BM25 initially, followed by transformer-based neural re-ranking models, specifically PubMedBERT and BioLinkBERT. To enhance the BM25 results they used the BGE-M3 model, and reciprocal rank fusion to combine model outputs. For phases A+ and B, their systems employed instruction-based transformer models such as llama-2, llama-3, Nous-Hermes2-Mixtral, and a BioASQ fine-tuned version of Gemma 2B for conditioned zero-shot answer generation. Specifically, they utilized the top-5 most relevant articles to generate an ideal answer and used relevant snippets in Phase B.

The CUHK-AIH team from The Chinese University of Hong Kong also participated in all phases. Their systems are based on a RAG  pipeline, with its base model being Llama2-chat-7B fine-tuned with LoRA using the training set. In Phase A, they first built indexes for all documents from PubMed Central using Pyserini to perform retrieval with BM25. In Phase A+, their systems further refined the retrieval by utilizing an ensemble retriever combining BM25 and vector similarity (bge-large-en embedding model). As for Phase B, the pipeline was similar to Phase A+, but the procedure was enhanced with the golden data.

The Gatech team from the Georgia Institute of Technology competed in all three phases. Their systems utilized a two-level information retrieval system and a QA system, based on pretrained LLMs including Mixtral, GPT-J, GPT-4 and llama-2, and prompt engineering. Specifically, their systems are based on LLM prompts with in-context few-shot examples to 1) parse the keywords in the given question to construct PubMed query and 2) solicit long and short answers for a question. Furthermore, they utilized post-processing techniques like resampling and malformed response detection to improve the performance.

The Fudan-Atypon team employed a two-stage IR model for phase A, similar to their previous work. In the first stage, they asked an LLM to extract words from the query, which were then used for query expansion.
For Phase A+ and Phase B, they applied prompt engineering and four LLMs (GPT3.5/4.0, ChatGLM, Spark) for exact answer generation. For the final answers, they combined the distinct results to ensure stable performance. As for the ideal answer, they used a scenario prompt, which asked LLMs to reply in the way of a student majoring in biology. They fine-tuned a LLama3-8B-instruct to answer the questions in precise sentences, which has been proven to be one of the best methods.

In phase A, the UNIPD team from the University of Padova participated with five systems. As a first step, their systems utilized GPT-3.5 and Gemini to generate pseudo-documents that contain relevant information to a question and extract biomedical entities from these pseudo-documents using NER tools. Then, their systems follow a two-stage retrieval. In the first-stage, they used BM25 with the original queries concatenated together with the biomedical entities. In the second stage, they employed a BiomedBERT cross-encoder re-ranker, which was trained on the combination of golden standard data, synthetic data, as well as the LLM-generated pseudo-documents.

The OPIX team participated with two systems in phase A. Their systems followed a two-stage retrieval. They initially employed sparse document retrieval, followed by re-ranking which calculated the cosine similarity between the dense query and document representations and combined it with the cumulative scores of the sparse retrieval. Furthermore, their sparse retriever was based on BM25 and a graph attention neural network (GAT) that shared birectional information with a BERT model to enhance the re-ranking step. They utilised also the domain-specific UMLS knowledge graph, linking the entities mentioned on the PubMed documents to entity nodes of the UMLS graph. Their systems take pairs of queries and relevant KG subgraphs as input and bidirectionally fuses information from both modalities creating dense representations for the query and the node entities of the graph.

Also, the HU team competed in phase A with five systems. Their systems focused on using BM25 for sparse first-stage retrieval and combining a variety of dense, neural models via rank fusion for second-stage re-ranking. The neural re-rankers consist of i) two distinct cross-encoders (MedCPT and E5), one trained on a pairwise loss and one trained on both token-level and document-level features, as well as ii) one LLM approach (GPT-3.5), generating synthetic queries from documents returned by the first-stage retrieval and comparing their similarities to the original test query.

In phase B, the NCU team from the National Central University participated with five systems. Their systems utlized GPT-4 and RAG techniques to improve the retrieval process. They 
employed prompt engineering to refine the input queries guiding GPT-4 and improving its output accuracy and relevance. RAG was integrated to retrieve relevant biomedical documents, which were then incorporated into the generation process. 
The UL team from the Uni de Lisboa competed with one system in phase B. Their system focused on enhancing LLMs with external biomedical data. Specifically, they utilized the Mistral-7B-Instruct v0.1 model and an iterative process of fine-tuning using manually curated biomedical data alongside open-source resources.    
The VCU team from Virginia Commonwealth University participated with four different systems in phase B. Their systems
are based on a zero-shot learning approach using generative LLMs, including Synthia-13B-GPTQ and llama-3. Their systems heavily relied on prompt engineering and answer processing.

\subsection{Task Synergy 12}

In the twelfth edition of BioASQ, four teams participated in the Synergy task (Synergy 12). These teams submitted results from 16 distinct systems. 
An overview of systems and approaches employed is provided in Table \ref{tab:syn}.

\begin{table}[!htb]
        \centering
         \caption{Systems and their approaches for task Synergy. Systems for which no description was available at the time of writing are omitted. }
        \begin{tabular}{M{0.2\linewidth}M{0.06\linewidth}M{0.7\linewidth}}\hline
        \textbf{System} & \textbf{Ref.} & \textbf{Approach} \\ \hline

        BSRC & \cite{Panou24} & open source LLMs, sparse/dense/hybrid retrieval \\\hline 
        UR & \cite{Ateia24clef} & GPT-3.5-turbo, GPT-4, query expansion\\\hline 
        Gatech & \cite{Zhou24clef} &  Mixtral, GPT-J, GPT-4, Llama2, resampling\\\hline 
        \end{tabular}
        \label{tab:syn}
\end{table}

In particular, the BSRC Alexander Fleming team participated with four systems. Similar to task b, their systems focused on LLMs with optimized prompts and majority voting.
Also, the UR team from the Universität Regensburg competed with two systems. Their systems employed 2-shot and zero-shot learning with GPT-3.5-turbo and GPT-4. Furthermore, their systems utilized handcrafted examples for the 2-shot learning as well as incorporated query expansion methods.
The Gatech team from the Georgia Institute of Technology participated in five systems. As with task b, their systems relied on pre-trained LLMs and prompt engineering.
More detailed descriptions for some of the systems are available at the proceedings of the workshop.

\subsection{Task MultiCardioNER}

31 teams registered for the MultiCardioNER task, out of which 7 teams submitted at least one run of their predictions. Specifically, 6 teams participated in the CardioDis subtrack, while 5 teams participated in the MultiDrug subtrack (with one of those teams participating only in the Spanish part). Overall, a total of 70 runs were submitted, with each team being allowed up to 5 runs per subtrack and language.

\begin{table}[!htb]
        \centering
         \caption{General overview of the approaches presented by participants for the MultiCardioNER task. ``*TEMIST corpora'' refers to the joint version of the DisTEMIST, SympTEMIST, ProcTEMIST and DrugTEMIST corpora.}
        \begin{tabular}{M{0.16\linewidth}M{0.04\linewidth}M{0.12\linewidth}M{0.63\linewidth}}\hline
        \textbf{Team} & \textbf{Ref.} & \textbf{Task} & \textbf{Approaches} \\ \hline
         BIT.UA & \cite{bitua} & CardioDis  & Ensemble of RoBERTa models with multi-head CRF and differences in the data used for training (only DisTEMIST or DisTEMIST + CardioCCC) \\
         \hline
        Data Science TUW & \cite{datasciencetuw} & CardioDis & Transformer-based models with different pretraining settings, data augmentation and window sliding \\
         & & MultiDrug & Multilingual and language-specific Transformers with different pretraining settings, data augmentation and window sliding \\
         \hline
         Enigma & \cite{enigma} & CardioDis & CLIN-X-ES model fine-tuned on the entire task data + custom clinical dataset \\
         & & MultiDrug & Multilingual and language-specific Transformers fine-tuned on the entire task data + custom drug dictionary \\
         \hline
         ICUE & \cite{icue} & MultiDrug & Multilingual and language-specific BERT models with re-training, post-processing rules + GPT 3.5 \\
         \hline
         NOVALINCS & \cite{novalincs} & CardioDis &  RoBERTa model fine-tuned on the standalone DisTEMIST corpus vs. joint *TEMIST corpora \\
         & & MultiDrug & RoBERTa model fine-tuned on the standalone DrugTEMIST corpus vs. joint *TEMIST corpora \\
         \hline
         PICUSLab & \cite{picuslab} & CardioDis & Ensemble of Transformer-based models trained on different datasets, including an augmented version of CardioCCC + post-processing via string matching \\
         \hline
         Siemens & \cite{siemens} & CardioDis & Fine-tuned general domain BERT model \\
         & & MultiDrug & Fine-tuned language-specific general domain BERT models \\
         \hline
         \end{tabular}
        \label{tab:multicardioner_sys}
\end{table}

Table \ref{tab:multicardioner_sys} gives an overview of the methodologies used by the participants in each of the sub-tasks. 
Following the trend of previous similar shared tasks \cite{medprocner,symptemist,meddoplace_sepln}, all participants used some variant of Transformers-based models, with RoBERTa \cite{roberta} models being the most popular. 
Other than that, ensembles were quite popular and provided good results (e.g. BIT.UA \cite{bitua}), as were the use of custom datasets and dictionaries (e.g. Enigma \cite{enigma}), data augmentation or window sliding (e.g. Data Science TUW \cite{datasciencetuw}).
It is also noteworthy the way in which the teams incorporated the cardiology-specific data, with some teams trying to mesh it into their training data in different ways (e.g. PICUSLab  \cite{picuslab}) and others using only the mixed-specialty training data (e.g. NOVALINCS \cite{novalincs}).
Finally, an interesting aspect of the MultiDrug subtrack is that, while the most common approach was to focus creating separate, language-specific models, there were some teams who tried to create purely multilingual models attempting to optimize the performance for all three languages at once, such as the ICUE team \cite{icue}.

\subsection{Task BioNNE}

26 teams registered for the BioNNE task in CodaLab, out of which 5 teams submitted at least one run of their predictions. Overall, a total of 155 runs were submitted. An overview of the approaches is provided in Table~\ref{tab:bionne}. Two systems for which no information was available at the time of writing are omitted. 

\begin{table}[!htb]
        \centering
         \caption{Overview of the approaches presented by participants for the BioNNE task. EN stands for the English-oriented and RU for the Russian-oriented tracks.}
        \begin{tabular}{M{0.17\linewidth}M{0.04\linewidth}M{0.23\linewidth}M{0.52\linewidth}}\hline
        \textbf{Team} & \textbf{Ref.} & \textbf{Track} & \textbf{Approaches} \\ \hline
         fulstock & - & Bilingual, EN, RU & BINDER, XLM-RoBERTa \\
         \hline
        wenxinzh & \cite{bionnewsenxin} & EN & Mixtral, spaCy NER, UMLS \\
         \hline
         hasin.rehana & \cite{bionnerehana} & Bilingual, EN, RU & PubMedBERT, SBERT-Large-NLU-RU, BERT-Base-Multilingual-uncased, UMLS \\
         \hline
         \end{tabular}
        \label{tab:bionne}
\end{table}

Team \textbf{fulstock} employed the BINDER model \cite{zhang-etal-2022-binder}, which uses XLM-RoBERTa \cite{XLMroberta} as its backbone. The team experimented with various entity type descriptions (prompts) for BINDER learning. These prompts included: keyword (name of the entity type), 2, 5, or 10 most frequent component words for entity type in the training data, contextual prompt (a sentence example with the target entity), and lexical prompt (a sentence example where the target entity is masked with the entity label) \cite{rozhkov2023prompts} . The model was trained over 64 epochs.

Team \textbf{wenxinzh} \cite{bionnewsenxin} combined a pretrained Mixtral model \cite{Mixtral} with a spaCy NER model trained on the BC5CDR corpus \cite{BC5CDR}. They also adapted and customized rules based on UMLS (Unified Medical Language System) semantic types. The system first utilizes Mixtral and en\_ner\_bc5cdr\_md to extract potential entities for each category from the text and then determines their final entity types by finding the associated UMLS semantic types. 

Team \textbf{hasin.rehana} \cite{bionnerehana} implemented the BIO-tagging scheme, applying six levels of BIO-tagging. They added six classification layers to the base model, each dedicated to outputting a specific level of NER tags. The original dataset's eight classes were expanded to 17 to accommodate the BIO-tagging scheme. For vocabulary expansion, they utilized the UMLS Metathesaurus to extract relevant and related concepts. For English NNER, they used the pre-trained PubMedBERT for contextualized word embeddings \cite{pubmedbert}; for Russian NNER, they employed a pre-trained SBERT-Large-NLU-RU model; and for Bilingual NNER, they utilized BERT-Base-Multilingual-uncased \cite{DBLP:journals/corr/abs-1810-04805}.

\section{Results}
\label{sec:results}

\subsection{Task 12b}
This section presents the evaluation measures and preliminary results for the task 12b. These results are preliminary, as the final results will be available after the manual assessment of the system responses by the BioASQ team of experts and the enrichment of the ground truth with potential additional relevant items, answer elements, and/or synonyms, which is still in progress.

\textbf{Phase A}: 
The Mean Average Precision (MAP) was used for evaluation on document retrieval. In particular, since BioASQ8, MAP calculation is based on a modified version of Average Precision (AP) that considers both the limit of 10 elements allowed per question in each submission and the actual number of golden elements that is often less than 10 in practice~\cite{nentidis2020overview}.  
For snippets, where a single ground-truth snippet may overlap with several submitted ones, the interpretation of MAP is less straightforward. 
Hence, since BioASQ9, we use the F-measure which is based on character overlaps\footnote{\url{http://participants-area.bioasq.org/Tasks/b/eval\_meas\_2022/}}~\cite{nentidis2021overview}.
Table \ref{tab:bA_res_doc} presents some indicative results for document retrieval in batch 1. The full 12b results for phase A are available online\footnote{\footnotesize \url{http://participants-area.bioasq.org/results/12b/phaseA/}}. 

\textbf{Phases A+ and B}: 
The official ranking for systems providing \textit{ideal answers} is based on manual scores assigned by the BioASQ team of experts that assesses each \textit{ideal answer} in the responses~\cite{balikas13}.  
The final position of systems providing \textit{exact answers} is based on their average ranking in the three question types where \textit{exact answers} are required, that is ``yes/no'', ``list'', and ``factoid''. Summary questions for which no \textit{exact answers} are submitted are not considered in this ranking.
In particular, the mean F1 measure is used for the ranking in list questions, the Mean Reciprocal Rank (MRR) is used for the ranking in factoid questions, and the F1 measure, macro-averaged over the classes of yes and no, is used for yes/no questions.
Tables~\ref{tab:bAp_res} and~\ref{tab:bB_res} present some indicative results on \textit{exact answer} extraction. The full 12b results for both phase A+\footnote{\footnotesize \url{http://participants-area.bioasq.org/results/12b/phaseAplus/}} and B\footnote{\footnotesize \url{http://participants-area.bioasq.org/results/12b/phaseB/}} are available online. 
\begin{table*}[!htbp]
\centering
\caption{Preliminary results for document retrieval in batch 1 of phase A of task 12b. 
Only the top-6 systems are presented, based on MAP.
}
\begin{tabular}{M{0.3\linewidth}M{0.13\linewidth}M{0.13\linewidth}M{0.13\linewidth}M{0.13\linewidth}M{0.12\linewidth}}\hline
\textbf{System} & \textbf{Mean Precision} & \textbf{Mean Recall} & \textbf{Mean F-measure} & \textbf{MAP} & \textbf{GMAP}  \\ \hline
bioinfo-4           & 0.1039         & 0.3124 & 0.1485    & \textbf{0.2067} & 0.0016 \\
bioinfo-3           & 0.1009         & 0.3047 & 0.1444    & 0.2024 & 0.0013 \\
bioinfo-1           & 0.1156         & 0.3171 & 0.1581    & 0.2018 & 0.0015 \\
bioinfo-2           & \textbf{0.1294}         & \textbf{0.3369} & \textbf{0.1728}    & 0.2006 & \textbf{0.0019} \\
bioinfo-0           & 0.1151         & 0.3055 & 0.1570    & 0.1800 & 0.0009 \\
dmiip2024\_3        & 0.0706         & 0.2514 & 0.1039    & 0.1612 & 0.0007 \\
\hline
\\
\end{tabular}
\label{tab:bA_res_doc}


\centering
\caption{Results for batch 1 for \textit{exact answers} in phase A+ of task 12b.
Only the top-6 systems based on Yes/No F1 are presented.}
\begin{tabular}
{M{0.205\linewidth}M{0.0853\linewidth}M{0.0852\linewidth}M{0.105\linewidth}M{0.11\linewidth}M{0.0852\linewidth}M{0.0852\linewidth}M{0.0852\linewidth}M{0.0852\linewidth}}
\hline

\textbf{System} & \multicolumn{2}{c}{\textbf{Yes/No}} & \multicolumn{3}{c}{\textbf{Factoid}} & \multicolumn{2}{c}{\textbf{List}} \\ 
\hline
& F1 & Acc. & Str. Acc. & Len. Acc. & MRR & Prec. & Rec. & F1 \\ \cline{2-9}    

UR-IW-3             & \textbf{0.9167}   &\textbf{ 0.920}   & 0.0952      & 0.0952       & 0.0952 & 0.4016     & 0.4778 & 0.4089    \\
Gatech comp...  & 0.8397   & 0.840   & 0.1429      & 0.1429       & 0.1429 & 0.4452     & 0.3415 & 0.3661    \\
GTBioASQsys3        & 0.8397   & 0.840   & 0.1429      & 0.1429       & 0.1429 & 0.2421     & 0.1765 & 0.1866    \\
UR-IW-4             & 0.8397   & 0.840   & 0.0476      & 0.0952       & 0.0714 & 0.3948     & 0.4063 & 0.3798    \\
UR-IW-2             & 0.8397   & 0.840   & 0.0952      & 0.0952       & 0.0952 & \textbf{0.5250}     & \textbf{0.4914} & \textbf{0.4808}    \\
UR-IW-5             & 0.7987   & 0.800   & 0.0952      & 0.0952       & 0.0952 & 0.4119     & 0.4182 & 0.3976    \\
\hline
\end{tabular}
\label{tab:bAp_res}

\centering
\caption{Results for batch 3 for \textit{exact answers} in phase B of task 12b.
Only the top-6 systems based on Yes/No F1 and the BioASQ Baseline are presented.}
\begin{tabular}
{M{0.205\linewidth}M{0.0852\linewidth}M{0.0852\linewidth}M{0.105\linewidth}M{0.11\linewidth}M{0.0852\linewidth}M{0.0852\linewidth}M{0.0852\linewidth}M{0.0852\linewidth}}
\hline

\textbf{System} & \multicolumn{2}{c}{\textbf{Yes/No}} & \multicolumn{3}{c}{\textbf{Factoid}} & \multicolumn{2}{c}{\textbf{List}} \\ 
\hline
& F1 & Acc. & Str. Acc. & Len. Acc. & MRR & Prec. & Rec. & F1 \\ \cline{2-9}    

mibi\_rag\_snippet     & \textbf{1.00}  &\textbf{1.00}  & 0.2308      & 0.2308       & 0.2308 & 0.4984     & 0.5157 & 0.5052    \\
RMC\_append\_sn... & \textbf{1.00}  &\textbf{1.00}  & 0.3077      & 0.3077       & 0.3077 & 0.4158     & 0.4475 & 0.3955    \\
Fleming-3              & \textbf{1.00}  &\textbf{1.00}  & 0.2308      & 0.2692       & 0.2404 & 0.5424     & 0.5532 & 0.5413    \\
IISR 4th submit        & \textbf{1.00}  &\textbf{1.00}  & 0.4231      & 0.4231       & 0.4231 & 0.5452     & 0.5187 & 0.5247    \\
dmiip2024\_2           & \textbf{1.00}  &\textbf{1.00}  & 0.1923      & 0.4231       & 0.2949 & 0.5114     & 0.5055 & 0.4715    \\
GTBioASQsys2           & 0.9577  & 0.9583  & 0.3846      & 0.3846       & 0.3846 & 0.5107     & 0.4774 & 0.4763    \\
BioASQ\_Baseline       & 0.4338  & 0.4583  & 0.0769      & 0.1538       & 0.1090 & 0.1999     & 0.2938 & 0.2094    \\
\hline
\end{tabular}

\label{tab:bB_res}
\end{table*}

\begin{figure*}[!htbp]
\centerline{\includegraphics[width=1\textwidth]{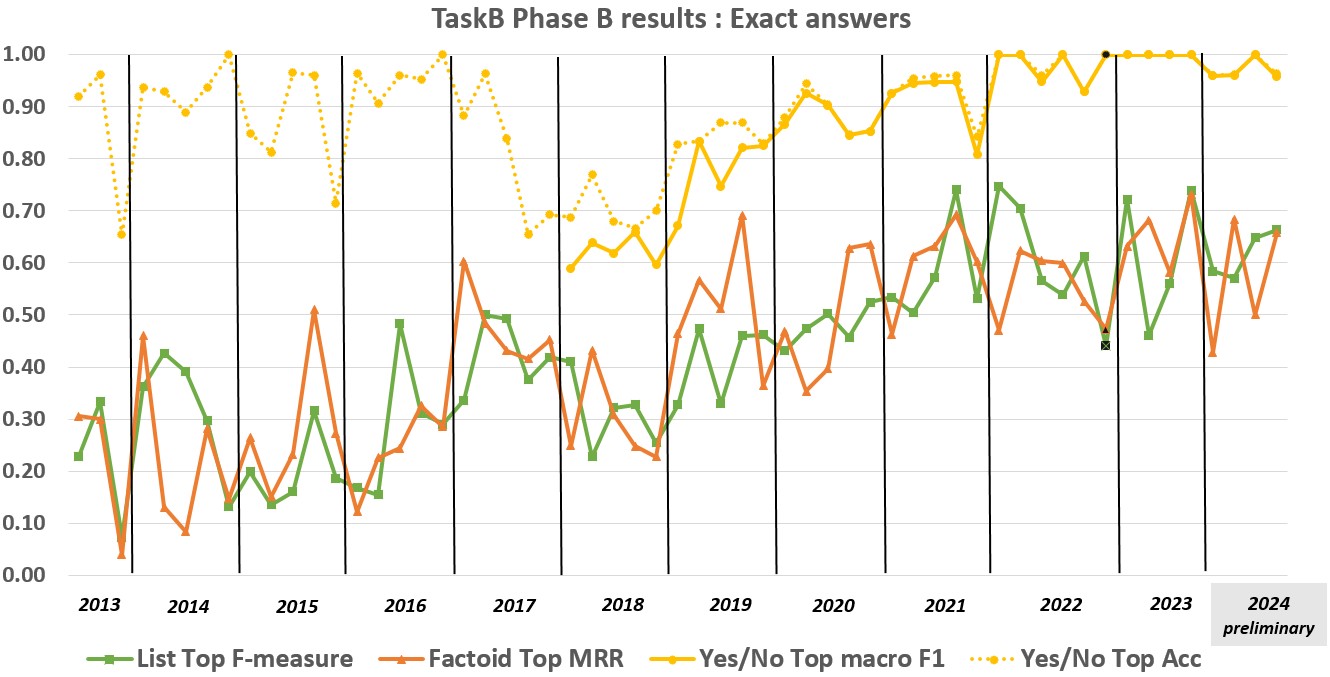}}
\caption{
The evaluation scores of the best-performing systems in task B, 
Phase B, for \textit{exact answers}, across the twelve years of BioASQ. Since BioASQ6, accuracy (Acc) was replaced by macro F1 as the official measure for Yes/No questions.
The black dots indicate an additional batch with questions from new experts~\cite{nentidis2021overview}.
}\label{fig:Exact}
\end{figure*}

The top performance of the participating systems in \textit{exact answer} generation for each type of question during the twelve years of BioASQ is presented in Figure {\ref{fig:Exact}}.
The preliminary results for task 12b, reveal that the participating systems keep achieving high scores in answering all types of questions, despite the addition of two new experts to the BioASQ team.
In batch 3, for instance, presented in Table \ref{tab:bB_res}, several systems manage to correctly answer literally all yes/no questions. High performance, beyond 0.95 in macro F1 is also observed for yes/no questions in the remaining batches.
More consistent performance is also observed in the preliminary results for list questions compared to the previous years, but there is still more room for improvement, as done for factoid questions where the performance across the batches fluctuates more.

\subsection{Task Synergy 12}

In task Synergy 12, no relevant material was initially available for new questions. For old questions, however, feedback from previous rounds was provided per question, that is the documents and snippets submitted by the participants with manual annotations of their relevance.
Hence, the documents and snippets of the feedback, that have already been assessed and released, were not considered valid for submission in the subsequent rounds. As in task 12b, the evaluation measures for document and snippet retrieval are MAP and F-measure respectively.

In addition, due to the developing nature of the topic, no answer is available for all of the open questions in each round. Therefore only the questions indicated as ``answer ready'' were evaluated for \textit{exact} and \textit{ideal answers} in each round.
Regarding the \textit{ideal answers}, the systems were ranked according to manual scores assigned to them by the BioASQ experts during the assessment of systems responses as in phase B of task B~\cite{balikas13}. 
As regards evaluation for the \textit{exact answers}, similarly to task 12b, the mean F1 measure, the Mean Reciprocal Rank (MRR), and the macro F1 measure are used for the ranking in list, factoid, and yes/no questions respectively.
Any \textit{exact} or \textit{ideal answer} that was assessed as ground-truth quality by the experts, was included in the feedback and provided to the participants before the next round.

\begin{table}
    \centering
     \caption{Results for document retrieval of the first round of the Synergy 12 task.}
    \begin{tabular}{M{0.3\linewidth}M{0.14\linewidth}M{0.12\linewidth}M{0.14\linewidth}M{0.12\linewidth}M{0.12\linewidth}}
    \hline
        \textbf{System} & \textbf{Mean precision} & \textbf{Mean Recall} & \textbf{Mean F-Measure} & \textbf{MAP} & \textbf{GMAP} \\ \hline
        dmiip3    & \textbf{0.4043}         & \textbf{0.4718} & \textbf{0.3558}    & \textbf{0.4636} & 0.1783 \\
        dmiip1    & 0.3899         & 0.4697 & 0.3472    & 0.4535 & 0.1697 \\
        dmiip4    & 0.3971         & 0.4674 & 0.3508    & 0.4493 & \textbf{0.1928} \\
        dmiip5    & 0.3971         & 0.4674 & 0.3508    & 0.4493 & \textbf{0.1928} \\
        dmiip2    & 0.3826         & 0.4690 & 0.3416    & 0.4427 & 0.1814 \\
        Fleming-4 & 0.3176         & 0.3385 & 0.2544    & 0.2342 & 0.0324 \\
        Fleming-2 & 0.2664         & 0.2524 & 0.2235    & 0.2152 & 0.0056 \\
        Fleming-1 & 0.2756         & 0.2476 & 0.2246    & 0.2110 & 0.0056 \\
        Fleming-3 & 0.2695         & 0.2696 & 0.2163    & 0.1985 & 0.0157 \\
        \hline 
    \end{tabular}
   \label{tab:synergy1-res}
\end{table}

Some indicative results for the Synergy task are presented in Table~\ref{tab:synergy1-res}.
The full 12b results are available online\footnote{\footnotesize \url{http://participants-area.bioasq.org/results/synergy\_v2024/}}. 
Overall, the collaboration between participating biomedical experts and question-answering systems allowed the progressive identification of relevant material and extraction of \textit{exact} and \textit{ideal answers} for several open questions for developing problems, such as Colorectal Cancer, pediatric sepsis, Duchenne Muscular Dystrophy, and COVID-19.   
In total, after the four rounds of Synergy 12, enough relevant material was identified to provide an answer to about 78\% of the questions. In addition, about 51\% of the questions had at least one \textit{ideal answer}, submitted by the systems, which was considered of ground-truth quality by the respective expert.

\subsection{Task MultiCardioNER}

All in all, the top scores for each subtrack were:

\begin{itemize}
  \item \textbf{CardioDis}. The team BIT.UA attained the top position with an ensemble of RoBERTa-based models (roberta-es-clinical-trials-ner) that also uses multi-head CRF. Their runs integrated the provided datasets in different ways, with the highest scores being achieved by the models that use both the DisTEMIST and CardioCCC data. Their best run achieved an F1-score of 0.8199 and a recall of 0.8243. The team with the next best F1-score (0.8049) is Enigma, who uses a CLIN-X-ES model also fine-tuned on the DisTEMIST and CardioCCC data. Interestingly, the team PICUSLab achieves the best precision (0.8886) by a wide margin combining the predictions of multiple models trained on different parts of the data (including an augmented version of the CardioCCC corpus) and then using string matching techniques to enhance the final predictions.
  \item \textbf{MultiDrug}. In Spanish, the best F1-score is achieved by the ICUE team (0.9277), who also achieved the best recall (0.9412). Meanwhile, in English and Italian the winner team is Enigma, with an F1-score of 0.9223 and 0.8842, respectively.
\end{itemize}

The results for the CardioDis subtrack are shown in in Table \ref{tab:multicardio_subtrack1}, while the results for the MultiDrug subtrack are presented in Table \ref{tab:multicardio_subtrack2-es} for Spanish, Table \ref{tab:multicardio_subtrack2-en} for English and Table \ref{tab:multicardio_subtrack2-it} for Italian. Due to space limitations, only the top-6 systems are presented, with the complete results being available in the MultiCardioNER overview paper \cite{multicardioner2024overview}.

\begin{table}[!ht]
    \centering
    \caption{Results of the MultiCardioNER CardioDis subtrack. Only the top-6 systems are presented. The best result is bolded, and the second-best is underlined.}
    \begin{tabular}{M{0.2\linewidth}M{0.4\linewidth}M{0.12\linewidth}M{0.12\linewidth}M{0.12\linewidth}}
    \hline
        \textbf{Team Name} & \textbf{Run name} & \textbf{Precision} & \textbf{Recall} & \textbf{F1} \\ 
        \hline
        BIT.UA & run1-all-full & 0.8155 & \textbf{0.8243} & \textbf{0.8199} \\ 
        BIT.UA & run0-top5-full & 0.811 & \underline{0.8181} & \underline{0.8145} \\ 
        Enigma & 3-system-CLIN-X-ES-pretrained & 0.8016 & 0.8082 & 0.8049 \\ 
        Enigma & 2-system-CLIN-X-ES-14 & 0.8052 & 0.8007 & 0.803 \\ 
        PICUSLab & aug\_fus\_sub2 & 0.7794 & 0.803 & 0.791 \\ 
        BIT.UA & run4-all & 0.7981 & 0.7827 & 0.7903 \\ 
        \hline
    \end{tabular}
    \label{tab:multicardio_subtrack1}

    \centering
    \caption{Results of the MultiCardioNER MultiDrug subtrack in Spanish. Only the top-6 systems are presented. The best result is bolded, and the second-best is underlined.}
    \begin{tabular}{M{0.2\linewidth}M{0.4\linewidth}M{0.12\linewidth}M{0.12\linewidth}M{0.12\linewidth}}
    \hline
        \textbf{Team Name} & \textbf{Run name} & \textbf{Precision} & \textbf{Recall} & \textbf{F1} \\ 
        \hline
        ICUE & run2\_single\_pp & 0.9146 & \textbf{0.9412} & \textbf{0.9277} \\ 
        ICUE & run4\_GPT\_translation & 0.9146 & 0.9412 & 0.9277 \\ 
        ICUE & run5\_GPT\_translation\_all & 0.9146 & 0.9412 & 0.9277 \\ 
        Enigma & 3-system-SpanishRoBERTa & 0.913 & \underline{0.9348} & \underline{0.9238} \\ 
        Enigma & 1-system-XLMR & 0.904 & 0.9208 & 0.9123 \\ 
        Enigma & 2-system-XLMR-filtering & \underline{0.9148} & 0.9005 & 0.9076 \\ 
        \hline
    \end{tabular}
    \label{tab:multicardio_subtrack2-es}

    \centering
    \caption{Results of the MultiCardioNER MultiDrug subtrack in English. Only the top-6 systems are presented. The best result is bolded, and the second-best is underlined.}
    \begin{tabular}{M{0.2\linewidth}M{0.4\linewidth}M{0.12\linewidth}M{0.12\linewidth}M{0.12\linewidth}}
    \hline
        \textbf{Team Name} & \textbf{Run name} & \textbf{Precision} & \textbf{Recall} & \textbf{F1} \\ 
        \hline
        Enigma & 3-system-BioLinkBERT & 0.8981 & \textbf{0.9477} & \textbf{0.9223} \\ 
        ICUE & run2\_single\_pp & \textbf{0.9086} & 0.9128 & \underline{0.9107} \\ 
        ICUE & run4\_GPT\_translation & 0.9086 & 0.9128 & 0.9107 \\ 
        Enigma & 1-system-XLMR & 0.8823 & 0.9233 & 0.9023 \\ 
        Enigma & 2-system-XLMR-filtering & \underline{0.9031} & 0.8989 & 0.901 \\ 
        Enigma & 5-system-XLMR-filtering-dict2 & 0.8698 & 0.9047 & 0.8869 \\ 
        \hline 
    \end{tabular}
    \label{tab:multicardio_subtrack2-en}

    \centering
    \caption{Results of the MultiCardioNER MultiDrug subtrack in Italian. Only the top-6 systems are presented. The best result is bolded, and the second-best is underlined.}
    \begin{tabular}{M{0.2\linewidth}M{0.4\linewidth}M{0.12\linewidth}M{0.12\linewidth}M{0.12\linewidth}}
    \hline
        \textbf{Team Name} & \textbf{Run name} & \textbf{Precision} & \textbf{Recall} & \textbf{F1} \\ 
        \hline
        Enigma & 1-system-XLMR & 0.884 & 0.8844 & \textbf{0.8842} \\ 
        Enigma & 3-system-Italian-Spanish-RoBERTa & 0.8723 & \underline{0.8956} & \underline{0.8838} \\ 
        Enigma & 2-system-XLMR-filtering & \underline{0.9016} & 0.8606 & 0.8806 \\ 
        Siemens & run1\_IMR & 0.8891 & 0.8689 & 0.8789 \\ 
        ICUE & run4\_GPT\_translation & \textbf{0.9114} & 0.8461 & 0.8776 \\ 
        ICUE & run5\_GPT\_translation\_all & 0.9114 & 0.8461 & 0.8776 \\ 
        \hline
    \end{tabular}
    \label{tab:multicardio_subtrack2-it}
\end{table}

In conclusion, the task's results are quite good and varied, with scores ranging from 0.9277 (by the ICUE team in the Spanish MultiDrug subtrack) to 0.2201 (by the DataScienceTUW team, who had some problems with the submission, also in the Spanish MultiDrug subtrack). Overall, the results for the MultiDrug subtrack are higher than those for the CardioDis subtrack, which was to be expected since drugs, as an entity type, are simpler and more straightforward than diseases.  

As stated earlier, in terms of methodology, there's a definite trend of using pre-trained Transformer-based systems (with a preference for RoBERTa models, perhaps due to their availability in Spanish), with most participants going beyond mere finetuning. Many of the presented runs incorporate new layers over their initial system results, be it an ensemble of multiple models and their predictions, multi-head CRFs, window sliding or using some kind of post-processing. 

However, what seems to really make a difference in MultiCardioNER is the use of the cardiology-specific data (the CardioCCC dataset), which is one of the shared task's main research points. All top-performing systems incorporate the released 258 documents from the CardioCCC corpus in some way. Meanwhile, participants that only use the DisTEMIST and DrugTEMIST corpora (which are made up of clinical case reports from varied clinical specialties) are able to achieve a really high precision but a much lower recall, thus achieving a not-so-high F1-score. This seems to indicate that, while these systems are able to retrieve many clinical entities correctly (i.e. high precision), they fail to recover those entities that are specific to the cardiology domain (i.e. low recall). 

Furthermore, comparing the results of the DisTEMIST shared task \cite{amiranda2022overview}, which also focused on diseases, with the CardioDis subtrack, shows an improvement in the overall results in this new task. All of this seems to point towards the importance of using data belonging to the clinical specialty that we plan to apply our systems to, even within domains that are already quite specific as is the clinical domain.
Still, it is true that, compared with DisTEMIST, this task offers a higher volume of data. While there seems to be a positive correlation with the use of domain-specific data, whether these improvements can actually be attributed to the domain adaptation aspect or to simply having more data remains to be seen and is a question for further research.

As for the multilingual aspect of the track, the results for all three languages are quite comparable, with Italian being somewhat below Spanish and English. This difference might be explained by the fact that, while there are many pre-trained models available in Spanish and English that were used by participants, this is not the case for Italian. In fact, the only Italian-specific models used were a version of BERT in Italian (that is, a general domain model) and BioBIT \cite{biobit}, a model that is specific to the biomedical domain trained on machine-translated PubMed abstracts. 

In contrast, participants were able to use a wider variety of models for English, such as BioLinkBERT \cite{linkbert} or SciBERT \cite{SciBERT}, and Spanish. An approach to solve this lack of clinical models in Italian followed by some participants was to further pre-train existing Spanish models using Italian and multilingual data, which made the Enigma team achieve the three top-scoring runs in the Italian track. Multilingual models such as mDeBERTa \cite{he2021deberta,he2021debertav3} were also used by participants.


\subsection{Task BioNNE}
The primary evaluation metric utilized in this study is the F$_{1}$-score, which is computed using the following formula: $F_1 = \frac{1}{n} \sum_{c \in C} F_{1_{rel_c}}$, where \( C \) represents the set of classes \{\texttt{FINDING}, \texttt{DISO}, \texttt{INJURY\_POISONING}, \texttt{PHYS}, \texttt{DEVICE}, \texttt{LABPROC}, \texttt{ANATOMY}, \texttt{CHEM}\}, \( n \) is the size of \( C \), and \( F_{1_{rel_c}} \) denotes the macro F$_1$-score averaged across all relevance classes.

\begin{table}[]
\caption{Results (F$_1$ scores on the test sets) of bilingual and monolingual subtasks. The best result in each task is bolded.} \label{tab:results}
\begin{center}
\begin{tabular}{|c|c|c|c|}
\hline
\textbf{Model} & \textbf{Both (Track 1) }& \textbf{English (Track 2)}  & \textbf{Russian (Track 3)} \\ \hline
fulstock  &  \textbf{0.7044} & \textbf{0.6181} & \textbf{0.6981} \\ 
hasin.rehana & 0.5053 & 0.5636 & 0.6007 \\ 
wenxinzh & -  & 0.348 & - \\ 
\hline
\end{tabular}
\end{center}
\end{table}

We summarized the performance of the above-mentioned teams in Table \ref{tab:results}. The fulstock team with the fine-tuned BINDER model achieved the highest F$_1$ scores across all tracks, with 0.704 for the bilingual track, 0.618 for the English-oriented track, and 0.698 for the Russian-oriented track.
In contrast, a pre-trained LLM, specifically the Mixtral model combined with the NER model for flat entities, achieved an F$_1$ score of 0.34797 for the English-oriented track. This score can be considered indicative of zero-shot evaluation, highlighting the limitations due to the absence of supervised training and the inadequacy of biomedical-specific training data in LLMs such as Mixtral. More results, along with baselines, are available in the BioNNE overview paper \cite{bionne}.

\section{Conclusions}
\label{sec:conclusion}

This paper provides an overview of the twelfth BioASQ challenge.
This year, BioASQ consisted of four tasks: (1) Task 12b on biomedical semantic question answering in English and (2) Synergy 12 on question answering for developing problems, both already established from previous BioASQ versions, (3) the new task MultiCardioNER on the automatic detection of disease and drug mentions on cardiology clinical case reports in Spanish, Italian, and English, and (4) the new task BIONNE on biomedical nested NER in English and Russian.

The preliminary results for task 12b reveal the high performance of the top participating systems, predominantly for yes/no answer generation, despite the extension of the expert team with two new experts. However, room for improvement is still available, particularly for factoid and list questions, where the performance is less consistent.
The results of the new Phase A+ also reveal that state-of-the-art QA approaches can achieve high performance, even without access to manually selected relevant material. Still, providing such material leads to answers of improved quality.  
This edition of the Synergy task as well, revealed that state-of-the-art systems, despite still having room for improvement, can be a useful tool for biomedical experts who need specialized information for addressing open questions in the context of several developing problems.

The new task MultiCardioNER presented two new challenging subtasks about annotations clinical case reports in Spanish, English, Italian with disease and drug mentions. 
Building on the work laid out in previous shared tasks like DisTEMIST \cite{amiranda2022overview}, this task introduces the nuance of creating clinical Named Entity Recognition systems specifically for the cardiology domain. In addition, it expands the range of the task beyond Spanish by introducing a subtrack that also involves English and Italian text. In order to do this, two new datasets are released: the DrugTEMIST corpus, that includes drug mentions in Spanish, English and Italian in a group of clinical case reports of varied medical specialties, and the CardioCCC corpus, a collection of cardiology clinical case reports with disease and drug annotations.
The results highlight the importance of having data specific to the language and specialty the systems are going to be applied in, even within domains that are already quite specific like the clinical one. 

The ever-increasing focus of participating systems on deep neural approaches and Large Language Models, already apparent in previous editions of the challenge, is also observed this year.
Most of the proposed approaches built on state-of-the-art neural architectures (BERT, PubMedBERT, BioBERT, BART etc.) adapted to the biomedical domain and specifically to the tasks of BioASQ. 
This year, in particular, several teams investigated approaches based on Generative Pre-trained Transformer (GPT) models and Retrieval Augmented Generation (RAG) for the BioASQ tasks. 

The BioNNE task centered on extracting the eight most common biomedical entities in Russian and English from PubMed abstracts while accommodating potential nested structures. The top-performing approach employed a bi-encoder framework that leverages contrastive learning to map text spans and entity types into a common vector representation space. The performance of pre-trained LLMs without fine-tuning exhibited significantly lower results, underscoring the necessity for specialized training data.

Overall, several systems managed competitive performance on the challenging tasks offered in BioASQ, as in previous versions of the challenge, and the top performing of them were able to improve over the state-of-the-art performance from previous years.
BioASQ keeps pushing the research frontier in biomedical semantic indexing and question answering for eleven years now, offering both well-established and new tasks. 
Aligned with the direction of extending beyond the English language and biomedical literature, which started with the task MESINESP \cite{luis2020overview} and continued consistently ever since, this year BioASQ was further extended with two new tasks, MultiCardioNER~\cite{multicardioner2024overview} and BioNNE~\cite{bionne}. 
In addition, this year we introduced a new phase in the QA task 12b (phase A+) allowing the assessment of systems that produce answers directly, without access to manually selected relevant material.
The future plans for the challenge include a further extension of the benchmark data for question answering through a community-driven process, extending the community of biomedical experts involved in the Synergy task, as well as extending the resources considered in the BioASQ tasks, both in terms of documents types, languages, and more focused sub-domains of biomedicine.

\section{Acknowledgments}
Google was a proud sponsor of the BioASQ Challenge in 2023. 
Ovid is also sponsoring this edition of BioASQ. 
The twelfth edition of BioASQ is also sponsored by Elsevier.
Atypon Systems Inc. is also sponsoring this edition of BioASQ. 
The MEDLINE/PubMed data resources considered in this work were accessed courtesy of the U.S. National Library of Medicine.
BioASQ is grateful to the CMU team for providing the \textit{exact answer} baselines for task 12b.
The MultiCardioNER track was funded by Spanish and European projects such as DataTools4Heart (Grant Agreement No. 101057849), AI4HF (Grant Agreement No. 101080430), BARITONE (Proyectos de Transición Ecológica y Transición Digital 2021. Expediente Nº TED2021-129974B-C21) and AI4ProfHealth (PID2020-119266RA-I00).
The work on the BioNNE task was supported by the Russian Science Foundation [grant number 23-11-00358].
%
%
%
\bibliographystyle{splncs04}
\bibliography{BioASQ11.bib}

\end{document}